\documentclass[lettersize,journal]{IEEEtran}
\usepackage{amsmath,amsfonts}
\usepackage{algorithmic}
\usepackage{algorithm}
\usepackage{array}
\usepackage[caption=false,font=normalsize,labelfont=sf,textfont=sf]{subfig}
\usepackage{textcomp}
\usepackage{stfloats}
\usepackage{url}
\usepackage{verbatim}
\usepackage{graphicx}
\usepackage{cite}
\usepackage{booktabs}
\usepackage{xcolor}
\usepackage{multirow}

\newcommand\blfootnote[1]{%
  \begingroup
  \renewcommand\thefootnote{}\footnote{#1}%
  \addtocounter{footnote}{-1}%
  \endgroup
}

\newcommand*{\our}{\texttt{RobustA}}

\begin{document}

\title{RobustA: \underline{Robust} \underline{A}nomaly Detection in Multimodal Data}

\author{Salem AlMarri$^{1 \dagger}$, Muhammad Irzam Liaqat$^{2 \dagger}$, Muhammad Zaigham Zaheer$^{1}$, Shah Nawaz$^{3}$, Karthik Nandakumar$^{1}$, Markus Schedl$^{3,4}$ \\
$^{1}$Mohamed Bin Zayed University of Artificial Intelligence,
$^{2}$IMT School for Advanced Studies, \\
$^{3}$Johannes Kepler University Linz,
$^{4}$Human-centered AI Group, AI Lab, Linz Institute of Technology
}



\maketitle

\begin{abstract}
\blfootnote{$\dagger$ Equal contribution}
In recent years, multimodal anomaly detection methods have demonstrated remarkable performance improvements over video-only models. However, real-world multimodal data is often corrupted due to unforeseen environmental distortions.
In this paper, we present the first-of-its-kind work that comprehensively investigates the adverse effects of corrupted modalities on multimodal anomaly detection task.  To streamline this work, we propose \our{}, a carefully curated evaluation dataset to systematically observe the impacts of audio and visual corruptions on the overall effectiveness of anomaly detection systems. Furthermore, we propose a multimodal anomaly detection method, which shows notable resilience against corrupted modalities. The proposed method learns a shared representation space for different modalities and employs a dynamic weighting scheme during inference based on the estimated level of corruption. Our work represents a significant step forward in enabling the real-world application of multimodal anomaly detection, addressing situations where the likely events of modality corruptions occur. The proposed evaluation dataset with corrupted modalities and respective extracted features will be made publicly available.
\end{abstract}



\section{Introduction}
\IEEEPARstart{V}{ideo} anomaly detection (VAD) is a challenging computer vision task with various real-world applications including surveillance, autonomous navigation, packaging, and biomedical imaging~\cite{liu2023generalized}.
Generally, VAD methods aim to predict high anomaly scores for the frames in a video that deviate significantly from the norm, where the application context determines the norm. For example, events such as shoplifting or violence can be considered anomalies in the CCTV surveillance context~\cite{wu2020not}.
Since it is laborious to obtain fine-grained (pixel-level or frame-level) labels of anomalies in videos, a weakly-supervised VAD (WS-VAD) setting that learns to detect anomalous frames using only video-level binary labels is typically used. 
In WS-VAD, a training video is labeled as normal if no anomalous event is present, whereas it is labeled as an anomaly if any anomalous event is present~\cite{sultani2018real,cho2023look,zaheer2022stabilizing}.   
In recent years, WS-VAD has gained considerable popularity and became a mainstream VAD paradigm~\cite{shou2018autoloc,narayan20193c,sultani2018real}.  
More recently, WS-VAD has been transformed into a multimodal learning task by including other informative modalities such as audio to discriminate and locate events better~\cite{wu2020not,yu2022modality}.
For instance, it is challenging to rely only on visual signals in shaky videos covering an explosion event. However, if accompanied by the audio modality, the event can be easily classified. 
Prior works have successfully shown the effectiveness of using video and audio pairs to improve the overall performance of WS-VAD~\cite{wu2020not,wu2022weakly,ghadiya2024cross}. 

\begin{figure}
    \centering
    \includegraphics[width=1\linewidth]{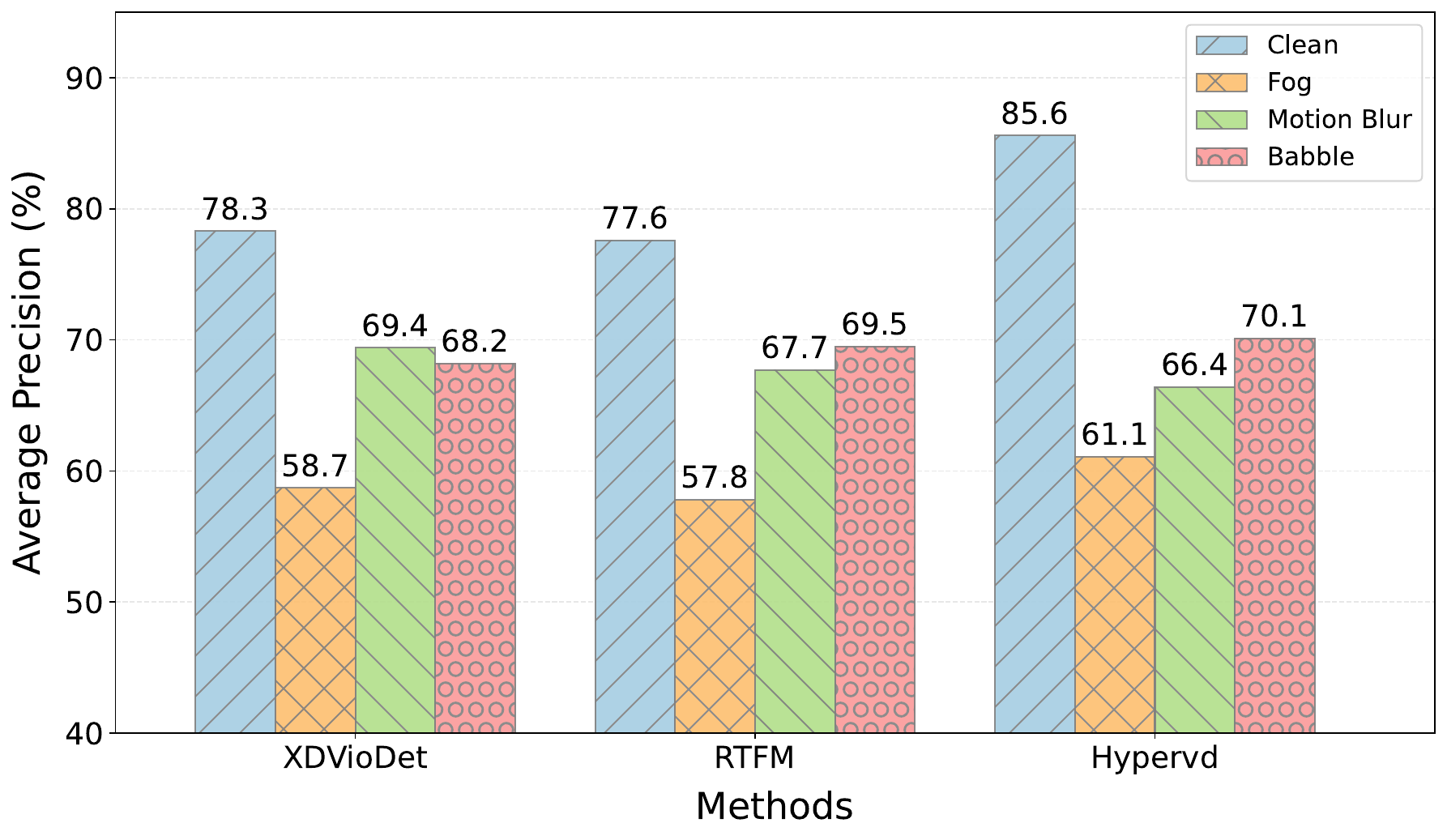}
    \caption{Existing multimodal anomaly detection approaches \cite{wu2020not,zhou2024learning,tian2021weakly} are not robust when subjected to corruptions in either modality, including vision (e.g., fog and motion blur) or audio (e.g., babble).}
    \label{fig:intro_fig}
\end{figure}


Despite their many advantages, deep neural network models are vulnerable to typical corruptions in the input data (e.g., degradations, distortions, and disturbances caused by weather changes, system errors, etc.) \cite{Josi_2023_WACV}. Though multimodal learning systems achieve better performance on multimodal data \cite{wu2024deep} compared to their unimodal counterparts, they are also susceptible to such vulnerabilities.
Multimodal VAD systems, in particular, are prone to such issues (see Fig. \ref{fig:intro_fig}), as they are often subjected to extreme environmental conditions \cite{yi2021benchmarking}.
For this reason, it is essential to design methods that can withstand corrupted data during deployment. While it may be possible to achieve this goal by including corrupted data during the training process, \textit{this work aims to evaluate the robustness of multimodal VAD systems trained only on clean data, when encountering corrupted data during deployment}. Existing literature lacks a detailed study and appropriate dataset that investigates the impact of corrupted modalities on WS-VAD and multimodal VAD systems.

In this work, we empirically investigate this problem with a proposed dataset, \our{}, specifically to study this issue. 
We observe that existing multimodal anomaly detection models deteriorate dramatically under corrupted data. For example, in Fig. \ref{fig:intro_fig}, the performance of XDVioDet~\cite{wu2020not} drops from $78.36$\% to $58.70$\% when exposed to fog in visual modality during testing.
To alleviate this issue, we propose a robust anomaly detection method that is resilient against corrupted modalities.
The proposed method independently maps multiple modalities into a shared representation space. This shared space can be particularly beneficial in scenarios where one modality is compromised. 
For example, if one modality becomes distorted, inaccurate, noisy, or is entirely missing, the shared space can leverage information from the remaining modalities to compensate for the loss. 
Results reported in Section~\ref{sec:experimental_results} suggest that our approach not only yields better multimodal performance but also demonstrates resilience against corruptions.

\noindent The key contributions of our work are as follows:
\begin{itemize}

    \item We present \our{}, a first-of-its-kind, carefully curated dataset comprising $8$ visual corruptions and $8$ audio corruptions, enabling evaluation of the multimodal anomaly detection methods against various types and levels of modality corruptions.

    \item We propose an anomaly detection approach that independently maps multiple modalities to learn representations in a shared space while dynamically adjusting the weights of individual modalities to reduce the impact of the compromised modality towards multimodal anomaly detection.
    
    \item  We study the plug-and-play nature of our proposed approach, demonstrating its resilience against compromised modalities when used in conjunction with existing approaches.
\end{itemize}


\section{Related Work}
\label{sec:related_work}
\subsection{Weakly Supervised Video Anomaly Detection and Multimodality}
Weakly Supervised Video Anomaly Detection (WS-VAD) refers to the process of identifying abnormal events in a given video while the training is carried out using only video-level binary labels.
The problem was first introduced by Sultani et al. \cite{sultani2018real} where the authors utilized multi-instance learning based ranking to carry out the overall training. The research was advanced by several researchers \cite{zaheer2020claws,zaheer2020self,zaheer2022generative}. However, most of these studies are focused on unimodal (video-only) training and testing.

More recently, anomaly detection task is transformed into multimodal learning where more than one modalities (usually vision and audio) are present during training and testing.
For example, XDVioDet~\cite{wu2020not} utilized a multi-branch neural network comprising holistic, localized, and score branches to effectively leverage multimodal feature representations. 
Similarly, HyperVD ~\cite{zhou2024learning} proposed a method using two graph-based hyperbolic convolutional networks for spatio-temporal feature extraction from multimodal input. While the multimodal anomaly detection task has gained popularity due to superior performance compared to the unimodal counterpart approaches, existing literature lacks rigorous studies understanding the adverse effects of compromised data on multimodal learning systems. As, CCTV surveillance cameras equipped with anomaly detection systems are highly prone to environmental corruptions, it is pertinent to explore this novel research direction.

\subsection{Addressing Compromised Modalities}
\noindent \textbf{Corrupted Modalities.} Data collected from diverse multiple sources may contain corrupted samples due to various factors, such as sensor failure, obstructions in the video stream, noise in audio signals, data storage errors, and many more. 
Recent years have seen an increased interest in investigating the vulnerability of deep models against modality corruptions~\cite{mitra2017robust,beemelmanns2024multicorrupt,kong2023robo3d,xie2023benchmarking,li2022deepfusion}.
For example, Hendrycks et al.~\cite{hendrycks2019benchmarking} built benchmarks to evaluate the performance of various Convolutional Network Networks on the image classification task. 
Similarly, Yi et al.~\cite{yi2021benchmarking} curated robust video classification benchmark to evaluate state-of-the-art Convolutional Networks and Transformers against video corruptions.
The growing interest has also broadened to multimodal learning where handling compromised modalities is critical to improve performance and robustness.
For example, Beemelmanns et al.~\cite{beemelmanns2024multicorrupt} introduces MultiCorrupt framework to evaluate the robustness of multimodal 3D object detectors against various corruption categories.
Similarly, Hong et al.~\cite{hong2023watch} proposed a multimodal input corruption modeling to develop robust audio-visual speech recognition models.
These studies demonstrate that the performance of unimodal or multimodal methods deteriorate dramatically when a modality is corrupted.

\noindent \textbf{Missing Modalities}
Multimodal learning has shown remarkable performance improvements over unimodal methods. However, such methods exhibit deteriorated performances if one or more modalities are missing~\cite{ma2021smil,ma2022multimodal,ganhor2024multimodal,liaqat2025chameleon}.
Considering the significance of multimodal learning, recent years have seen an increased interest in studies addressing missing modalities for various tasks including classification, such as ~\cite{lee2023multimodal,john2025multimodal,lin2023missmodal}.

\noindent While these methods have notably enhanced model robustness in addressing compromised modalities across various application areas, to the best of our knowledge, there remains a gap in the literature for the multimodal anomaly detection task. In this work, we investigated the robustness of multimodal anomaly detection under a carefully crafted dataset of compromised modalities. 
Moreover, we propose a robust approach that maintains superior performance when faced with extreme missing and corrupted modalities.

\begin{figure*}[t]
    \centering
    \includegraphics[width=1.05\textwidth]{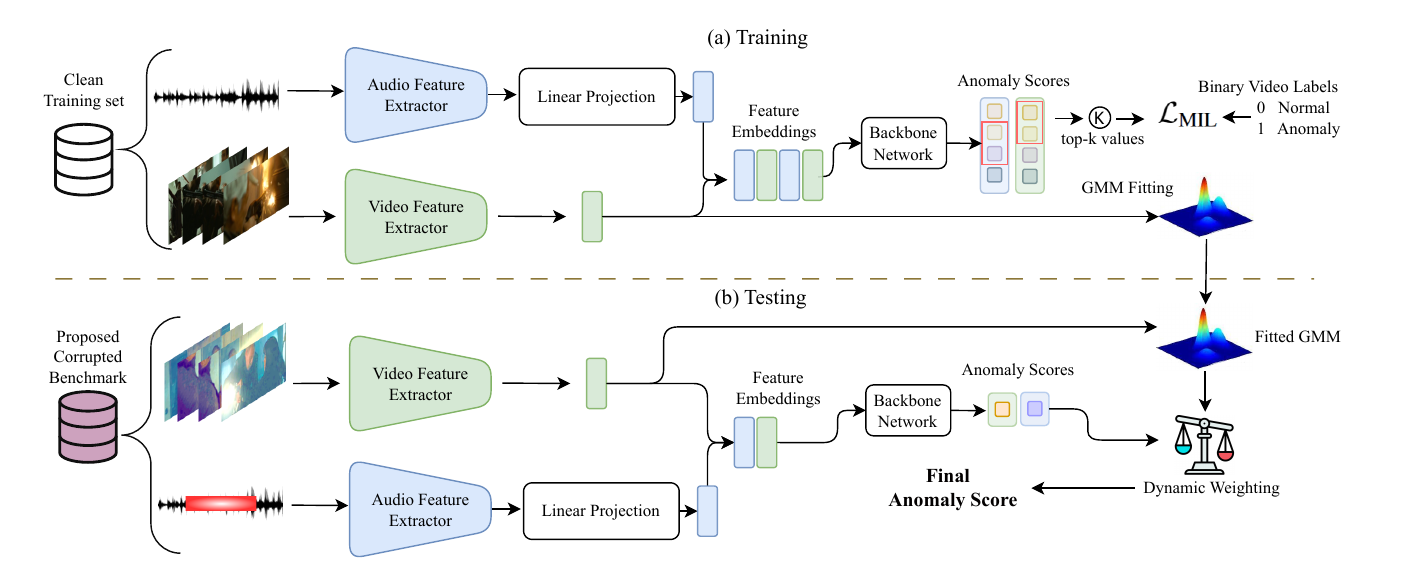} 
    \caption{Architecture of our approach. Modality-specific features are extracted using pre-trained audio and visual encoders. A linear projection is used to match the feature dimensions. Modality embeddings are independently mapped to learn representations in a shared space. During inference, the shared learning space helps mitigate the adverse effects of modality corruptions. Moreover, a dynamic weighting scheme is utilized to adjust the weights of the corrupted modality for better anomaly detection.}
    \label{fig:methodology}
\end{figure*}

\section{Methodology}
\subsection{Background and Overview}
Existing methods generally concatenate audio and visual embeddings to learn fused representations for the multimodal anomaly detection task~\cite{wu2020not, wu2022weakly,zhou2024learning,al2024collaborative}. 
While effective under modality-complete settings, these methods rely heavily on the complete availability of modalities and suffer dramatic performance deterioration when a modality is corrupted or missing. Since surveillance data may involve corruptions due to weather conditions, connection latency, broken/tempered equipment, etc., a lack of robustness against such scenarios may deter the deployment of these methods in real-world applications.
In this work, we develop an approach to \textit{learn multimodal representations by mapping audio and visual modalities independently into a common feature space}.
This enables the model to maintain \textit{robustness when a modality is compromised} by leveraging a shared representation of cross-modal interactions, effectively compensating for missing or corrupted modality data. 
The proposed approach is illustrated in Fig.~\ref{fig:methodology} and the details are discussed next.

\subsection{Preliminaries}
\label{sec:feature_extraction}

Given a dataset of \( n \) videos, each video \( V \) is preprocessed into \( m \) non-overlapping audio and visual segments and passed to a pre-trained feature extractor.
The audio and video feature extractors generate the feature embeddings \( E^A \in \mathbb{R}^{m \times d_A} \) and \( E^V \in \mathbb{R}^{m \times d_V} \) for each video, respectively with the feature dimensions are denoted as \( d_A \) and \( d_V \).
The class label \( y \in \{1,0\} \) indicates the presence ($1$) or absence ($0$) of anomalous events in the video. The goal is to learn an anomaly detector $\mathcal{A}_{\theta}( E^*)$ that takes a feature vector (audio or visual) as input and predicts an anomaly score in the range of $0$ and $1$.

\subsection{Learning a Shared Representation Space}
\label{sec:proposed_method}
Prior multimodal anomaly detection methods rely on concatenation of audio and visual representations and learning the anomaly detector based on the concatenated representation. Given the audio and visual feature embeddings \( E^A \) and \( E^V \) for \( n \) training videos in a dataset, Eq.~\ref{eq:baseline_loss} outlines the traditional fusion-based learning approach utilizing Multi-Instance Learning (MIL) loss~\cite{wu2020not}: 

\begin{equation}
\mathcal{L}_{\text{total}} = \sum_{i=1}^{n} \mathcal{L}_{\text{MIL}}(\mathcal{A}_{\theta}(E^A || E^V), y_i),
\label{eq:baseline_loss}
\end{equation}

\noindent where \( E^A || E^V \) is the fused multimodal representation obtained by concatenation of \( E^V \) and  \( E^A \), allowing the architecture to capture the inter-modality relationship. Consequently, the success of these methods is highly dependent on the availability of both modalities during inference.
To alleviate this dependence on modality completeness, we propose a method that learns multimodal representations by mapping each modality independently into a shared space.

%
Specifically, we map individual modality independently into a common representation space, enabling the model to  
capture inter- and intra-modality relationships through cross-modal data interactions. 
As a result, when a modality is compromised, the model compensates by leveraging the shared representation space.
Thus, Eq.~\ref{eq:baseline_loss} needs to be modified as follows:     

\begin{equation}
\mathcal{L}_{\text{total}} = \sum_{i=1}^{2 n} \mathcal{L}_{\text{MIL}}(\mathcal{A}_{\theta}(E^*_i), y_i) 
\label{eq:our_loss}
\end{equation}

\noindent where \( E^*\) is randomly sampled from either \( E^V \) or \( E^A \).  
However, in existing methods \cite{wu2020not}, depending on the choice of feature extractor, the output dimensions of \( E^V \) and \( E^A \) are usually inconsistent. To address this issue, we introduce a linear projection module $\mathcal{P}_{\phi}$ that transforms \( E^A \) to a higher dimensional space, matching the dimension of \( E^V \). Thus, \( E^*\) in Eq. (\ref{eq:our_loss}) will be randomly sampled from either \( E^V \) or \(\mathcal{P}_{\phi}(E^A) \). The parameters $\theta$ of the anomaly detector and $\phi$ of the projection module are learned by minimizing the total loss $\mathcal{L}_{\text{total}}$. Intuitively, the proposed approach attempts to learn a single modality-agnostic anomaly detector, which implicitly forces the two modalities to share a common representation space.

\subsection{Dynamic Weighting During Inference}
\label{sec:dynamic_weighting}

As the anomaly detector is agnostic to the input modality, it is possible to obtain an anomaly score using both audio and visual features during inference. Hence, a strategy to optimally combine individual anomaly scores is needed to achieve good multimodal performance. A straightforward approach would be to compute the weighted average of the anomaly scores produced by the anomaly detector $\mathcal{A}$ for each modality in a late fusion manner. For instance, the final anomaly score $S$ for a given video  segment based on audio-visual features can be obtained as:

\begin{equation}
S = \lambda_A \mathcal{A}(\overline{E}^{A})+ \lambda_V \mathcal{A}(\overline{E}^V),
\label{eq:navie_approach}
\end{equation}

\noindent  where $\overline{E}^{A}$ and $\overline{E}^{V}$ are the audio and visual feature embeddings of the test sample, respectively, which may also contain a compromised modality. Here, $\lambda_A$ and $\lambda_V$ are the weights assigned to the audio and visual modalities, respectively, and these weights are linearly normalized such that they add up to $1$. In a naive approach, $\lambda_A$ and $\lambda_V$ can be set to $0.5$ giving equal weightage to both modalities. While, as shown in our experiments, this approach works reasonably well due to the shared representation space mitigating the impacts of an individual compromised modality, a more sophisticated approach may be required to handle extreme corruption in one of the modalities. 

Since the training data does not have prior information about the compromised modality (since training is performed only on clean data), we devise a dynamic weighting scheme based on the clean data distribution. Specifically, we fit a Gaussian Mixture Model (GMM) to the clean (non-corrupted) training visual features and evaluate if the test data matches with this clean data distribution. Let $\mathcal{G^V}$ denote a $K$-component GMM with parameters $\{\pi_k, \boldsymbol{\mu}_k, \boldsymbol{\Sigma}_k\}_{k=1}^K$ representing the clean distribution of visual features, where $\pi_k$, $\boldsymbol{\mu}_k$, and $\boldsymbol{\Sigma}_k$ denote the mixture probability, mean, and covariance matrix, respectively, of the $k$-th Gaussian component. 

The negative log-likelihood $\ell$ of the test features $(\overline{E}^V)$ is then computed using the fitted GMM as follows: 

\begin{equation}
\ell\bigl(\overline{E}^V\bigr)
\;=\;
-\log
\Biggl[
\sum_{k=1}^{k}
\pi_{k}\,\mathcal{N}\Bigl(\overline{E}^V\,\big|\,
\boldsymbol{\mu}_{k}, \boldsymbol{\Sigma}_{k}\Bigr)
\Biggr]
\label{eq:gmm_loglik}
\end{equation}

\noindent where $\mathcal{N}$ denotes the likelihood of observing the feature vector $\overline{E}^V$ under the 
$k^{th}$ Gaussian component of the mixture model. A higher $\ell$ value indicates that the sample is more likely to be compromised, while a lower value of $\ell$ suggests the sample is clean. Finally, $\lambda_V$ is computed using a sigmoid function as:


\begin{equation}
\lambda_V 
\;=\;
\frac{0.5}{\,1 \;+\;
\exp\!\Bigl(\,c
\bigl[\,
\ell\bigl(\overline{E}^V\bigr) 
+x_0\bigr]\Bigr)
},
\label{eq:loglik_to_weight}
\end{equation}

\noindent where $c$ and $x_0$ are the scale and shift hyperparameters that shape the sigmoid function, respectively. The above equation maps the negative log-likelihood $\ell\left(\bar{E}^V\right)$ to a value of $\lambda_V$ between $0$ and $0.5$. Similarly, the value of $\lambda_A$ can also be computed and the two weights can be linearly normalized so that they sum to $1$. Extensive ablation studies on the proposed dynamic weighting scheme are provided in Section \ref{sec:ablation_dynamic}, which demonstrate the effectiveness of our approach. When one of the modalities is completely missing, its corresponding weight is set to zero because the anomaly score for that modality cannot be computed.

\begin{figure*}[!tbp]
    \centering
    \includegraphics[width=0.99\textwidth]{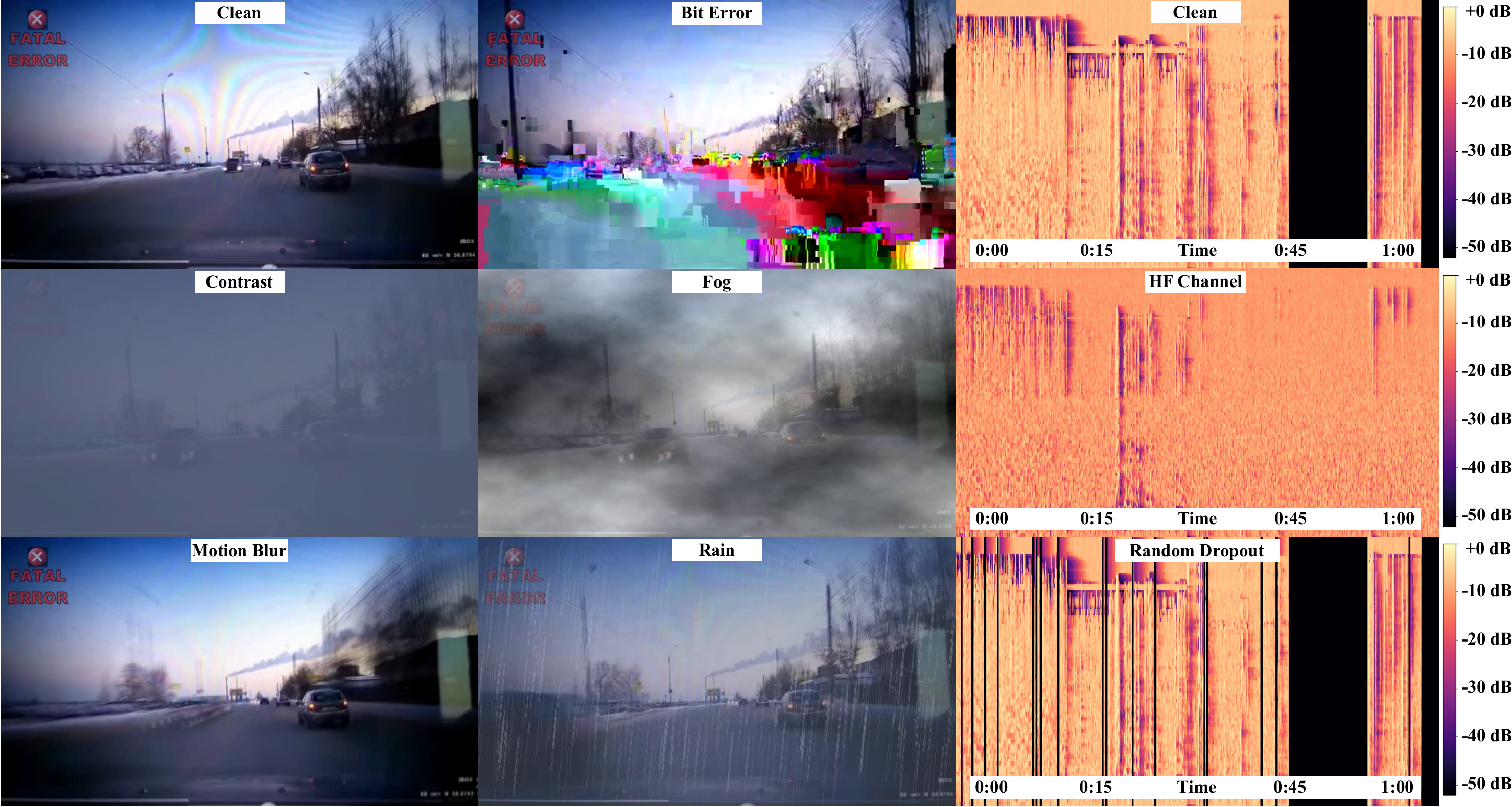} 
    \caption{Examples of a few visual and audio corruptions demonstrating the challenging scenarios presented in our proposed benchmark. 
    }
    \label{fig:rgb_corruption_visuals}
\end{figure*}

\section{Modality Corruptions}
\label{sec:benchmark}
In this work, we propose a carefully curated dataset, \our{}, to evaluate the critical impacts of audio/visual corruptions on the overall performance of multimodal anomaly detection systems. To the best of our knowledge, \our{} is the first rigorous attempt to study anomaly detection in such scenarios. We leverage XD-Violence multimodal anomaly detection dataset as the base to create several corruption scenarios, by varying the type and severity level, for both audio and visual modalities.
We choose these modalities because of their globally standard availability in the existing real-world CCTV systems, while other modalities, such as text, lack practical application.
Examples of some visual corruptions and audio corruption mel spectograms \cite{lostanlen2018per,mcfee2015librosa} are provided in Figure \ref{fig:rgb_corruption_visuals}.

\noindent \textbf{Types of Corruptions in Visual Modality.} We consider the following corruptions in the visual modality: Bit error, brightness, contrast, fog, rain, motion blur, saturation, and shot noise. Details about each corruption and its implementation are provided in the Supplementary.

\noindent \textbf{Types of Corruptions in Audio Modality.} We consider the following corruptions in the audio modality: Overlay, bit error,  pitch shift, random dropout, and reverb. 
Details about each corruption and the implementation details are provided in the Supplementary.


\noindent \textbf{Number of Corrupted Samples.} We consider varying numbers of corrupted modality samples. In particular, we consider 0\%, 10\%, 30\%, 50\%, 70\%, 90\%, and 100\% of samples in each modality as corrupted.

\setlength{\fboxsep}{0pt}  
\setlength{\fboxrule}{0pt} 

\begin{table*}[ht] 
\caption{Comparison of our approach and baseline under various corruption types
and corruption levels (percentages). AP(\%) is used as the evaluation metric (higher is better). Bold represents best performance in each comparison.}
\centering
\scriptsize

\scalebox{0.95}{
\begin{tabular}{ll|c|c|c|c|c|c|c}
\toprule
\multicolumn{2}{c|}{\textbf{Corruption Levels}} 
& \textbf{0\%} & \textbf{10\%} & \textbf{30\%} & \textbf{50\%} & \textbf{70\%} & \textbf{90\%} & \textbf{100\%} \\
\midrule
\textbf{Corruption} & \textbf{Modality}
& \cite{wu2020not}\quad{\textcolor{black}{Ours}}
& \cite{wu2020not}\quad{\textcolor{black}{Ours}}
& \cite{wu2020not}\quad{\textcolor{black}{Ours}}
& \cite{wu2020not}\quad{\textcolor{black}{Ours}}
& \cite{wu2020not}\quad{\textcolor{black}{Ours}}
& \cite{wu2020not}\quad{\textcolor{black}{Ours}}
& \cite{wu2020not}\quad{\textcolor{black}{Ours}} \\
\midrule

\textbf{Bit Error} & \textbf{Visual}
  & \underline{78.36} \quad{\textcolor{black}{\textbf{80.00}}}
  & \underline{77.41} \quad{\textcolor{black}{\textbf{79.19}}}
  & \underline{72.57}\quad{\textcolor{black}{\textbf{76.67}}}
  & \underline{72.48}\quad{\textcolor{black}{\textbf{76.35}}}
  & \underline{68.21}\quad{\textcolor{black}{\textbf{74.70}}}
  & \underline{66.61}\quad{\textcolor{black}{\textbf{74.78}}}
  & \underline{65.55}\quad{\textcolor{black}{\textbf{74.45}}} \\

\textbf{Brightness} & \textbf{Visual}
  & \underline{78.36}\quad{\textcolor{black}{\textbf{80.00}}}
  & \underline{77.39}\quad{\textcolor{black}{\textbf{79.30}}}
  & \underline{76.35}\quad{\textcolor{black}{\textbf{77.74}}}
  & \underline{75.54}\quad{\textcolor{black}{\textbf{77.02}}}
  & \underline{75.06}\quad{\textcolor{black}{\textbf{76.23}}}
  & \underline{74.34}\quad{\textcolor{black}{\textbf{75.42}}}
  & \underline{73.76}\quad{\textcolor{black}{\textbf{74.72}}} \\

\textbf{Contrast} & \textbf{Visual}
  & \underline{78.36}\quad{\textcolor{black}{\textbf{80.00}}}
  & \underline{77.08}\quad{\textcolor{black}{\textbf{78.67}}}
  & \underline{69.84}\quad{\textcolor{black}{\textbf{75.75}}}
  & \underline{66.82}\quad{\textcolor{black}{\textbf{74.44}}}
  & \underline{59.85}\quad{\textcolor{black}{\textbf{71.86}}}
  & \underline{54.54}\quad{\textcolor{black}{\textbf{70.35}}}
  & \underline{52.47}\quad{\textcolor{black}{\textbf{69.21}}} \\

\textbf{Fog} & \textbf{Visual}
  & \underline{78.36}\quad{\textcolor{black}{\textbf{80.00}}}
  & \underline{76.45}\quad{\textcolor{black}{\textbf{78.05}}}
  & \underline{69.21}\quad{\textcolor{black}{\textbf{73.47}}}
  & \underline{68.87}\quad{\textcolor{black}{\textbf{72.87}}}
  & \underline{62.32}\quad{\textcolor{black}{\textbf{69.68}}}
  & \underline{60.48}\quad{\textcolor{black}{\textbf{69.26}}}
  & \underline{58.70}\quad{\textcolor{black}{\textbf{67.97}}} \\

\textbf{Rain} & \textbf{Visual}
  & \underline{78.36}\quad{\textcolor{black}{\textbf{80.00}}}
  & \underline{77.73}\quad{\textcolor{black}{\textbf{79.52}}}
  & \underline{74.27}\quad{\textcolor{black}{\textbf{77.66}}}
  & \underline{72.88}\quad{\textcolor{black}{\textbf{77.16}}}
  & \underline{70.94}\quad{\textcolor{black}{\textbf{75.54}}}
  & \underline{69.43}\quad{\textcolor{black}{\textbf{74.56}}}
  & \underline{68.69}\quad{\textcolor{black}{\textbf{73.89}}} \\

\textbf{Motion Blur} & \textbf{Visual}
  & \underline{78.36}\quad{\textcolor{black}{\textbf{80.00}}}
  & \underline{77.22}\quad{\textcolor{black}{\textbf{78.73}}}
  & \underline{74.45}\quad{\textcolor{black}{\textbf{77.02}}}
  & \underline{74.03}\quad{\textcolor{black}{\textbf{76.46}}}
  & \underline{71.43}\quad{\textcolor{black}{\textbf{74.55}}}
  & \underline{70.28}\quad{\textcolor{black}{\textbf{73.80}}}
  & \underline{69.46}\quad{\textcolor{black}{\textbf{73.08}}} \\

\textbf{Saturate} & \textbf{Visual}
  & \underline{78.36}\quad{\textcolor{black}{\textbf{80.00}}}
  & \underline{77.02}\quad{\textcolor{black}{\textbf{78.84}}}
  & \underline{73.64}\quad{\textcolor{black}{\textbf{77.14}}}
  & \underline{69.94}\quad{\textcolor{black}{\textbf{76.10}}}
  & \underline{68.20}\quad{\textcolor{black}{\textbf{75.19}}}
  & \underline{65.57}\quad{\textcolor{black}{\textbf{74.26}}}
  & \underline{65.23}\quad{\textcolor{black}{\textbf{73.86}}} \\

\textbf{Shot Noise} & \textbf{Visual}
  & \underline{78.36}\quad{\textcolor{black}{\textbf{80.00}}}
  & \underline{77.69}\quad{\textcolor{black}{\textbf{79.10}}}
  & \underline{74.65}\quad{\textcolor{black}{\textbf{76.08}}}
  & \underline{74.60}\quad{\textcolor{black}{\textbf{75.62}}}
  & \underline{72.35}\quad{\textcolor{black}{\textbf{73.62}}}
  & \underline{71.40}\quad{\textcolor{black}{\textbf{72.72}}}
  & \underline{70.99}\quad{\textcolor{black}{\textbf{71.68}}} \\

\textbf{Average} & \textbf{Visual}
  & \underline{78.36}\quad{\textcolor{black}{\textbf{80.00}}}
  & \underline{77.25}\quad{\textcolor{black}{\textbf{78.93}}}
  & \underline{73.12}\quad{\textcolor{black}{\textbf{76.44}}}
  & \underline{71.90}\quad{\textcolor{black}{\textbf{75.75}}}
  & \underline{68.54}\quad{\textcolor{black}{\textbf{73.92}}}
  & \underline{66.58}\quad{\textcolor{black}{\textbf{73.14}}}
  & \underline{65.61}\quad{\textcolor{black}{\textbf{72.36}}} \\
\midrule

\textbf{Babble} & \textbf{Audio}
  & \underline{78.36}\quad{\textcolor{black}{\textbf{80.00}}}
  & \underline{77.16}\quad{\textcolor{black}{\textbf{79.06}}}
  & \underline{75.18}\quad{\textcolor{black}{\textbf{77.64}}}
  & \underline{72.65}\quad{\textcolor{black}{\textbf{74.36}}}
  & \underline{70.94}\quad{\textcolor{black}{\textbf{73.29}}}
  & \underline{68.57}\quad{\textcolor{black}{\textbf{70.96}}}
  & \underline{68.21}\quad{\textcolor{black}{\textbf{70.45}}} \\

\textbf{Bitrate} & \textbf{Audio}
  & \underline{78.36}\quad{\textcolor{black}{\textbf{80.00}}}
  & \underline{77.03}\quad{\textcolor{black}{\textbf{78.45}}}
  & \underline{75.60}\quad{\textcolor{black}{\textbf{76.66}}}
  & {\textcolor{black}{\textbf{72.76}}}\quad{\underline{72.37}} 
  & {\textcolor{black}{\textbf{71.12}}}\quad{\underline{70.21}} 
  & {\textcolor{black}{\textbf{68.67}}}\quad{\underline{65.48}} 
  & {\textcolor{black}{\textbf{68.31}}}\quad{\underline{65.11}} \\

\textbf{HF Channel} & \textbf{Audio}
  & \underline{}78.36\quad{\textcolor{black}{\textbf{80.00}}}
  & \underline{76.37}\quad{\textcolor{black}{\textbf{79.01}}}
  & \underline{73.73}\quad{\textcolor{black}{\textbf{77.77}}}
  & \underline{69.36}\quad{\textcolor{black}{\textbf{73.24}}}
  & \underline{67.59}\quad{\textcolor{black}{\textbf{72.09}}}
  & \underline{65.35}\quad{\textcolor{black}{\textbf{69.78}}}
  & \underline{65.56}\quad{\textcolor{black}{\textbf{69.47}}} \\

\textbf{Pink} & \textbf{Audio}
  & \underline{78.36}\quad{\textcolor{black}{\textbf{80.00}}}
  & \underline{76.32}\quad{\textcolor{black}{\textbf{79.02}}}
  & \underline{73.45}\quad{\textcolor{black}{\textbf{77.94}}}
  & \underline{68.97}\quad{\textcolor{black}{\textbf{73.63}}}
  & \underline{67.30}\quad{\textcolor{black}{\textbf{72.67}}}
  & \underline{64.99}\quad{\textcolor{black}{\textbf{70.85}}}
  & \underline{65.36}\quad{\textcolor{black}{\textbf{70.60}}} \\

\textbf{Pitch Shift} & \textbf{Audio}
  & \underline{78.36}\quad{\textcolor{black}{\textbf{80.00}}}
  & \underline{76.24}\quad{\textcolor{black}{\textbf{78.87}}}
  & \underline{73.49}\quad{\textcolor{black}{\textbf{77.62}}}
  & \underline{69.53}\quad{\textcolor{black}{\textbf{72.83}}}
  & \underline{67.01}\quad{\textcolor{black}{\textbf{71.23}}}
  & \underline{63.66}\quad{\textcolor{black}{\textbf{66.91}}}
  & \underline{63.07}\quad{\textcolor{black}{\textbf{66.39}}} \\

\textbf{Random Dropout} & \textbf{Audio}
  & \underline{78.36}\quad{\textcolor{black}{\textbf{80.00}}}
  & \underline{76.45}\quad{\textcolor{black}{\textbf{78.93}}}
  & \underline{74.22}\quad{\textcolor{black}{\textbf{77.70}}}
  & \underline{69.80}\quad{\textcolor{black}{\textbf{73.05}}}
  & \underline{67.79}\quad{\textcolor{black}{\textbf{71.81}}}
  & \underline{64.01}\quad{\textcolor{black}{\textbf{66.92}}}
  & \underline{63.53}\quad{\textcolor{black}{\textbf{66.48}}} \\

\textbf{Reverb} & \textbf{Audio}
  & \underline{78.36}\quad{\textcolor{black}{\textbf{80.00}}}
  & \underline{76.65}\quad{\textcolor{black}{\textbf{79.01}}}
  & \underline{75.02}\quad{\textcolor{black}{\textbf{77.78}}}
  & \underline{71.99}\quad{\textcolor{black}{\textbf{73.23}}}
  & \underline{70.01}\quad{\textcolor{black}{\textbf{71.90}}}
  & \underline{67.01}\quad{\textcolor{black}{\textbf{70.02}}}
  & \underline{66.84}\quad{\textcolor{black}{\textbf{69.73}}} \\

\textbf{White} & \textbf{Audio}
  & \underline{78.36}\quad{\textcolor{black}{\textbf{80.00}}}
  & \underline{76.48}\quad{\textcolor{black}{\textbf{78.95}}}
  & \underline{73.69}\quad{\textcolor{black}{\textbf{77.81}}}
  & \underline{69.24}\quad{\textcolor{black}{\textbf{73.48}}}
  & \underline{67.88}\quad{\textcolor{black}{\textbf{72.36}}}
  & \underline{66.02}\quad{\textcolor{black}{\textbf{70.43}}}
  & \underline{66.60}\quad{\textcolor{black}{\textbf{70.21}}} \\

\textbf{Average} & \textbf{Audio}
  & \underline{78.36}\quad{\textcolor{black}{\textbf{80.00}}}
  & \underline{76.59}\quad{\textcolor{black}{\textbf{78.91}}}
  & \underline{74.30}\quad{\textcolor{black}{\textbf{77.74}}}
  & \underline{70.41}\quad{\textcolor{black}{\textbf{73.27}}}
  & \underline{68.71}\quad{\textcolor{black}{\textbf{71.94}}}
  & \underline{66.04}\quad{\textcolor{black}{\textbf{68.92}}}
  & \underline{65.94}\quad{\textcolor{black}{\textbf{68.56}}} \\
\midrule

\textbf{Missing Visual} & \textbf{Visual}
  & \underline{78.30}\quad{\textcolor{black}{\textbf{80.00}}}
  & \underline{75.10}\quad{\textcolor{black}{\textbf{76.94}}}
  & \underline{69.50}\quad{\textcolor{black}{\textbf{71.87}}}
  & \underline{63.10}\quad{\textcolor{black}{\textbf{71.18}}}
  & \underline{58.20}\quad{\textcolor{black}{\textbf{68.61}}}
  & \underline{52.10}\quad{\textcolor{black}{\textbf{68.00}}}
  & \underline{49.50}\quad{\textcolor{black}{\textbf{67.15}}} \\

\textbf{Missing Audio} & \textbf{Audio}
  & \underline{78.36}\quad{\textcolor{black}{\textbf{80.00}}}
  & \underline{77.30}\quad{\textcolor{black}{\textbf{78.86}}}
  & \underline{75.20}\quad{\textcolor{black}{\textbf{77.96}}}
  & \underline{73.90}\quad{\textcolor{black}{\textbf{74.75}}}
  & \underline{72.50}\quad{\textcolor{black}{\textbf{73.91}}}
  & \underline{70.00}\quad{\textcolor{black}{\textbf{72.74}}}
  & \underline{69.10}\quad{\textcolor{black}{\textbf{72.55}}} \\
\midrule

\textbf{Mixed Corruptions} & \textbf{Visual}
  & \underline{78.36}\quad{\textcolor{black}{\textbf{80.00}}}
  & \underline{77.70}\quad{\textcolor{black}{\textbf{79.54}}}
  & \underline{73.93}\quad{\textcolor{black}{\textbf{78.15}}}
  & \underline{73.75}\quad{\textcolor{black}{\textbf{78.65}}}
  & \underline{72.75}\quad{\textcolor{black}{\textbf{77.35}}}
  & \underline{72.09}\quad{\textcolor{black}{\textbf{77.15}}}
  & \underline{71.45}\quad{\textcolor{black}{\textbf{76.27}}} \\

\textbf{Mixed Corruptions} & \textbf{Audio}
  & \underline{78.36}\quad{\textcolor{black}{\textbf{80.00}}}
  & \underline{76.94}\quad{\textcolor{black}{\textbf{79.09}}}
  & \underline{74.77}\quad{\textcolor{black}{\textbf{77.62}}}
  & \underline{72.40}\quad{\textcolor{black}{\textbf{74.13}}}
  & \underline{70.48}\quad{\textcolor{black}{\textbf{72.99}}}
  & \underline{67.71}\quad{\textcolor{black}{\textbf{70.50}}}
  & \underline{67.14}\quad{\textcolor{black}{\textbf{69.97}}} \\

\midrule
\textbf{Motion Blur$+$Babble} & \textbf{Audio$+$Visual}
  & \underline{78.36}\quad{\textcolor{black}{\textbf{80.00}}}
  & \underline{75.71}\quad{\textcolor{black}{\textbf{77.59}}}
  & \underline{70.91}\quad{\textcolor{black}{\textbf{73.54}}}
  & \underline{67.19}\quad{\textcolor{black}{\textbf{69.03}}}
  & \underline{61.94}\quad{\textcolor{black}{\textbf{64.22}}}
  & \underline{56.82}\quad{\textcolor{black}{\textbf{60.13}}}
  & \underline{54.66}\quad{\textcolor{black}{\textbf{58.04}}} \\

\textbf{Brightness$+$Rand. Dropout} & \textbf{Audio$+$Visual}
  & \underline{78.36}\quad{\textcolor{black}{\textbf{80.00}}}
  & \underline{75.42}\quad{\textcolor{black}{\textbf{78.08}}}
  & \underline{71.66}\quad{\textcolor{black}{\textbf{74.49}}}
  & \underline{65.20}\quad{\textcolor{black}{\textbf{67.83}}}
  & \underline{61.62}\quad{\textcolor{black}{\textbf{65.11}}}
  & \underline{56.05}\quad{\textcolor{black}{\textbf{56.94}}}
  & \underline{54.19}\quad{\textcolor{black}{\textbf{55.12}}} \\

\textbf{Bit Error$+$HF-Channel Noise} & \textbf{Audio$+$Visual}
  & \underline{78.36}\quad{\textcolor{black}{\textbf{80.00}}}
  & \underline{74.56}\quad{\textcolor{black}{\textbf{78.07}}}
  & \underline{69.43}\quad{\textcolor{black}{\textbf{74.29}}}
  & \underline{62.89}\quad{\textcolor{black}{\textbf{70.27}}}
  & \underline{55.53}\quad{\textcolor{black}{\textbf{65.87}}}
  & \underline{47.73}\quad{\textcolor{black}{\textbf{60.60}}}
  & \underline{45.50}\quad{\textcolor{black}{\textbf{59.47}}} \\

\bottomrule
\end{tabular}
}

\label{tab:baseline_proposed_modality}
\end{table*}

\section{Experiments}

\subsection{Implementation Details}
\textbf{Feature Extraction}: For feature extraction of visual modality, we employ ResNet-I3D model pretrained on Kinetics-$400$ dataset.
Following prior work~\cite{tian2021weakly}, we perform $10$-crop augmentation by using a $16$ frame sliding window with a sample rate of $24$ FPS. 
The audio embeddings are extracted by using the VGGish network \cite{gemmeke2017audio}  pre-trained on large-scale YouTube dataset. 
Moreover, the extraction is performed by pre-processing the audio signal into \(960\) ms  of overlapping segments and setting \(96 \times 64 \) bin size for the mel-spectrogram.

\noindent\textbf{Training Details:} We trained the proposed architecture on  Nvidia RTX $A100$ for $50$ epochs using Adam optimizer with initial learning rate of \( 10^{-5}\) and dynamic adjustment using multi-step cosine scheduler. We use a batch size of $640$ and weight decay of $0.00001$. 

\noindent\textbf{Backbone Network:} We adopt XDVioDet~\cite{wu2020not} as the primary backbone network in our experiments (also referred to as baseline hereafter) and provide extensive comparisons with it. 
In addition, to explore the general applicability of our proposed approach, we provide additional experiments on two other models as backbone networks, including RTFM~\cite{tian2021weakly} and HyperVD~\cite{zhou2024learning}. Nevertheless, unless specified otherwise, the results are reported with XDVioDet as the primary backbone network. 

\noindent\textbf{Dataset:} We use XD-Violence multimodal anomaly detection dataset~\cite{wu2020not} as a stepping stone for our dataset. 
More specifically, we apply corruptions on the test set of the XD-Violence dataset based on the details provided in Section~\ref{sec:benchmark} to create \our{}.
Similar to the original dataset, the train and test splits include $3,954$ and $800$ anomalous and normal videos, with varying levels of corruptions based on the experiment.
Following prior works~\cite{wu2020not, zhou2024learning}, we use Average Precision (AP) as the evaluation metric.

\subsection{Experimental Results}
\label{sec:experimental_results}

\subsubsection{Robustness Against Compromised Modalities}

Table \ref{tab:baseline_proposed_modality} provides extensive performance evaluation of our method and the baseline \cite{wu2020not} under various types of audio and visual corruptions. It may be noted that the training set remains clean, as our goal is to achieve robustness against corruptions when trained with clean data.

\noindent\textbf{Visual Corruptions:} During testing, various types of corruptions, such as bit error, fog, rain etc., are present in the visual modality (as outlined in Section \ref{sec:benchmark}).
We evaluate the performance of our method against the baseline on varying levels of corruption, which reflect the proportion of corrupted video in the test set. A corruption level of $0$\% means no corrupted video, while a level of $100$\% indicates that all test videos are corrupted.
The results (Table \ref{tab:baseline_proposed_modality}) demonstrate that our method outperformed the baseline against various types and levels of corruptions. 
Notably, for extreme cases of corruptions such as contrast, our method achieves performance scores of $80.00$\% and $69.21$\%, whereas the baseline attains scores of $78.36$\% and $52.47$\% under $0$\% and $100$\% of corrupted videos.
Although the performance of our method and the baseline method is comparable under $0$\% corruption, our method notably outperforms it under $100$\% corruption, highlighting its robustness against visual modality corruption.

\noindent \textbf{Audio Corruptions.} During testing, various types of corruptions, such as babble, bitrate, and hfchannel, are present in the audio modality (as described in Section \ref{sec:benchmark}). Consistent with the trends observed in the visual corruption section, we find similar patterns for the audio modality.
The results highlight that our method is resilient against various levels of audio corruptions. 

\noindent \textbf{Missing Modalities.} While corruptions interrupt the overall quality of a modality, some information may still be retained, which might be useful for the network to perform the predictions. To take the challenge up a notch, in this section, we consider the missing modality setting in which a modality is entirely removed. This may be considered as an extreme case where no information regarding the corrupted modality is available to the network.
Although the performances of our method ($80.00$\%) and the baseline ($78.30$\%) are comparable when modalities are completely available, our method notably outperforms the baseline method under missing modalities. Notably, our method achieves performance of $67.15$\% and $72.55$\%, whereas the baseline method attains scores of $49.50$\% and $69.10$\% when removing visual and audio modality, respectively, highlighting the robustness of our approach.

\noindent \textbf{Mixed Corruptions.}
As real-world environments may have several distortions present, we also explore the effects of the modalities compromised by mixed corruptions. To carry out this experiment, for each level of corruption, it is ensured that the test set contains all corruption types accounting for the total percentage of the corruption level. Performance trends similar to the previously discussed experiments are observed, where our approach outperforms the baseline in all compared cases, signifying the importance of our method against modality corruptions in real-world settings.

\noindent \textbf{Audio and Visual Corruptions.}
Since real-world environments often contain multiple distortions at the same time, we investigate their impact on both modalities.
Consistent with prior experiments, the corruptions cause significant drops in the performance of the baseline, whereas our approach demonstrates better resilience.

\begin{table}[t]
\caption{Comparison of baseline vs.\ proposed when one modality is completely missing and the other is corrupted. Higher accuracy is better.}
\centering
\scriptsize
\resizebox{0.99\linewidth}{!}{
\begin{tabular}{l l c c c c}
\toprule
\multicolumn{2}{c}{\textbf{Corruption Levels}} & \textbf{0\%} & \textbf{30\%} & \textbf{70\%} & \textbf{100\%} \\
\midrule
\textbf{Corruption} & \textbf{Mod} & \cite{wu2020not}\quad\textbf{Ours} & \cite{wu2020not}\quad\textbf{Ours} & \cite{wu2020not}\quad\textbf{Ours} & \cite{wu2020not}\quad\textbf{Ours} \\
\midrule
\textbf{bit\_error}   & V & \underline{69.17}\quad\textbf{73.82} & \underline{63.76}\quad\textbf{67.31} & \underline{61.57}\quad\textbf{64.70} & \underline{59.23}\quad\textbf{63.42} \\
\textbf{brightness}   & V & \underline{69.17}\quad\textbf{73.82} & \underline{65.64}\quad\textbf{70.27} & \underline{63.82}\quad\textbf{68.76} & \underline{60.84}\quad\textbf{66.45} \\
\textbf{contrast}     & V & \underline{69.17}\quad\textbf{73.82} & \underline{59.51}\quad\textbf{65.79} & \underline{48.12}\quad\textbf{55.83} & \underline{38.65}\quad\textbf{49.07} \\
\textbf{fog}          & V & \underline{69.17}\quad\textbf{73.82} & \underline{56.97}\quad\textbf{60.76} & \underline{52.48}\quad\textbf{54.49} & \underline{50.25}\quad\textbf{50.41} \\
\textbf{rain}         & V & \underline{69.17}\quad\textbf{73.82} & \underline{63.89}\quad\textbf{69.15} & \underline{59.60}\quad\textbf{64.49} & \underline{58.11}\quad\textbf{60.33} \\
\textbf{motion\_blur} & V & \underline{69.17}\quad\textbf{73.82} & \underline{63.21}\quad\textbf{68.83} & \underline{58.21}\quad\textbf{64.79} & \underline{54.61}\quad\textbf{62.29} \\
\textbf{saturate}     & V & \underline{69.17}\quad\textbf{73.82} & \underline{63.27}\quad\textbf{68.94} & \underline{57.61}\quad\textbf{64.61} & \underline{54.16}\quad\textbf{62.22} \\
\textbf{shot\_noise}  & V & \underline{69.17}\quad\textbf{73.82} & \underline{65.11}\quad\textbf{67.97} & \textbf{64.65}\quad\underline{63.23} & \textbf{65.44}\quad\underline{59.28} \\
\midrule
\textbf{average}      & V & \underline{69.17}\quad\textbf{73.82} & \underline{62.67}\quad\textbf{67.38} & \underline{58.26}\quad\textbf{62.61} & \underline{55.16}\quad\textbf{59.18} \\
\midrule
\textbf{babble}       & A & \underline{49.50}\quad\textbf{67.65} & \underline{41.42}\quad\textbf{59.27} & \underline{32.94}\quad\textbf{42.12} & \underline{28.47}\quad\textbf{31.47} \\
\textbf{bitrate}      & A & \underline{49.50}\quad\textbf{67.65} & \underline{43.10}\quad\textbf{56.95} & \underline{34.36}\quad\textbf{40.57} & \underline{28.74}\quad\textbf{32.11} \\
\textbf{hfchannel}    & A & \underline{49.50}\quad\textbf{67.65} & \underline{40.92}\quad\textbf{58.91} & \underline{29.05}\quad\textbf{36.21} & \textbf{24.55}\quad\underline{24.01} \\
\textbf{pink}         & A & \underline{49.50}\quad\textbf{67.65} & \underline{41.92}\quad\textbf{59.12} & \underline{28.86}\quad\textbf{36.80} & \textbf{25.93}\quad\underline{24.57} \\
\textbf{pitch\_shift} & A & \underline{49.50}\quad\textbf{67.65} & \underline{36.68}\quad\textbf{57.96} & \underline{26.11}\quad\textbf{34.90} & \underline{22.23}\quad\textbf{23.27} \\
\textbf{dropout}      & A & \underline{49.50}\quad\textbf{67.65} & \underline{38.74}\quad\textbf{58.46} & \underline{27.53}\quad\textbf{37.90} & \underline{23.62}\quad\textbf{27.20} \\
\textbf{reverb}       & A & \underline{49.50}\quad\textbf{67.65} & \underline{40.07}\quad\textbf{58.47} & \underline{30.83}\quad\textbf{37.22} & \underline{26.09}\quad\textbf{26.39} \\
\textbf{white}        & A & \underline{49.50}\quad\textbf{67.65} & \underline{42.38}\quad\textbf{58.90} & \underline{29.77}\quad\textbf{35.85} & \textbf{24.40}\quad\underline{23.40} \\
\midrule
\textbf{average}      & A & \underline{49.50}\quad\textbf{67.65} & \underline{40.65}\quad\textbf{58.51} & \underline{29.93}\quad\textbf{37.70} & \underline{25.50}\quad\textbf{26.55} \\

\bottomrule
\end{tabular}
}
\label{tab:unimodal}
\end{table}

\begin{figure*}
    \centering
    \includegraphics[width=0.99\textwidth]{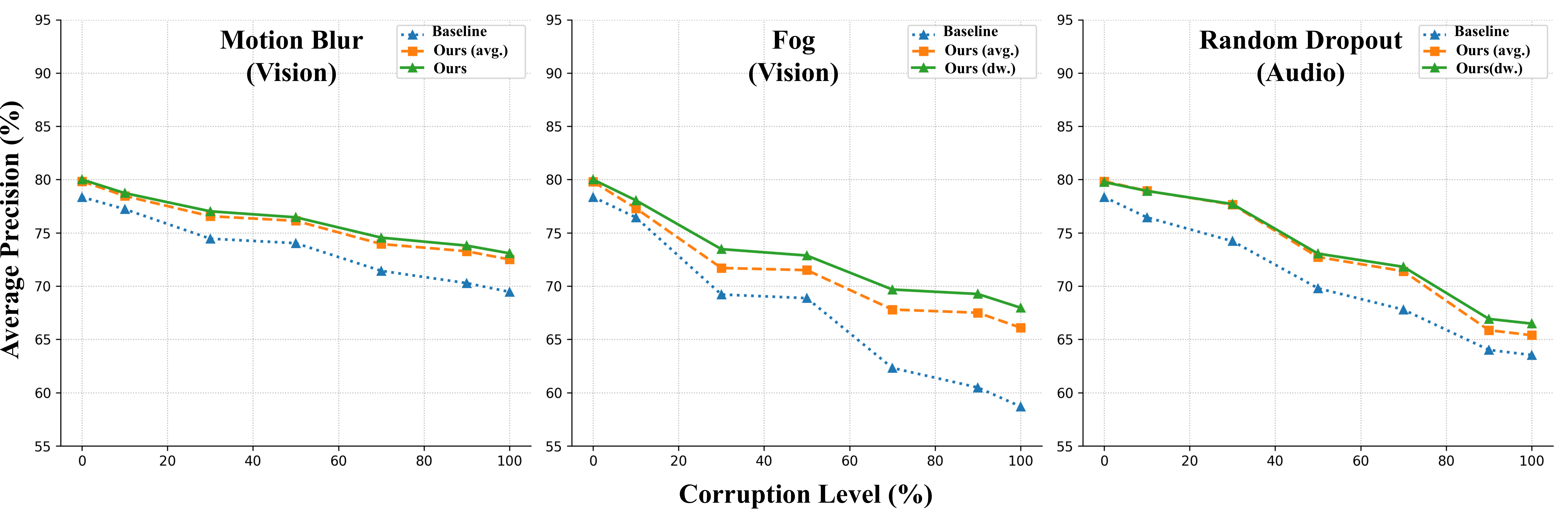} 
    \caption{Comparison of weighting schemes (average and dynamic) with baseline concatenation approach~\cite{wu2020not} on two visual and one audio corruption.}
    \label{fig:comparison_bs_prop_propdw}
\end{figure*}

\subsubsection{Unimodal Testing with Corruptions}
In a series of experiments, we explore a use case on our method in which one modality is completely missing while the other modality is corrupted. The intuition behind this experiment is to observe the importance of each modality for training and testing a robust multimodal anomaly detection system. Table \ref{tab:unimodal} summarizes the results obtained during these experiments. It may be seen that the video modality is notably dominant than the audio modality. In the case of $0$\% corruption cases, the presence of video modality outperforms the presence of audio modality (AP of $73.82$\% vs. $67.65$\%). When the noise is added to the visual modality, a notable performance drop is observed in each corruption case (an average drop in AP from $73.80$\% to $59.18$\%). However, the performance drop increases dramatically when the system is exposed to $100$\% missing visual modality while adding noise to the audio modality (an average drop in AP from $67.65$\% to $26.55$\%). This series of experiments highlights that while audio modality plays an important role in the overall performance, the visual modality is generally more informative, and the performance deteriorates dramatically if the visual modality is compromised.

\begin{table}[t]
    \caption{Plug-and-play performance of our approach for mixed audio and visual distortions at 0\% and 100\% levels of corruption on multiple baselines including XDVioDet, HyperVD, and RTFM. Bold: best performance.}
    \centering
    \footnotesize
    \begin{tabular}{l cc cc}
        \toprule
        \textbf{Corrupted Modality} & \multicolumn{2}{c}{\textbf{Visual}} & \multicolumn{2}{c}{\textbf{Audio}} \\
        \cmidrule(lr){2-3} \cmidrule(lr){4-5}
         & \textbf{0\%} & \textbf{100\%} & \textbf{0\%} & \textbf{100\%} \\

        \midrule
        XDVioDet (Baseline) & \underline{78.36} & \underline{71.45} & \underline{78.36} & \underline{67.14} \\
        XDVioDet (Ours)   & \textbf{80.00 }& \textbf{76.27 }& \textbf{80.00} & \textbf{69.97} \\
        \midrule
        HyperVD (Baseline) & 85.60 & 73.28 & 85.60 & 76.90 \\
        HyperVD (Ours)     & \textbf{85.90} & \textbf{77.29} & \textbf{85.90} & \textbf{78.18} \\
        \midrule
        RTFM (Baseline)     & \textbf{77.66} & \underline{70.43} & \textbf{77.66} & \textbf{77.44} \\
        RTFM (Ours)     & \underline{77.30} & \textbf{72.67 }& \underline{77.30} & \underline{74.21} \\
        \bottomrule
    \end{tabular}

    \label{tab:corruption_results}
\end{table}

\section{Analysis and Discussion}

\subsection{Importance of Dynamic Weighting}
\label{sec:ablation_dynamic}
During training, our method processes each modality independently, while at test time, we employ a dynamic weighting strategy (Section \ref{sec:dynamic_weighting}) to effectively combine the predictions of individual modalities for multimodal performance.
In this section, we compare the empirical performance of three approaches: the naive averaging approach (as chalked in Eq. \ref{eq:navie_approach}), dynamic weighting (as chalked in Eq. \ref{eq:loglik_to_weight}), and the concatenation-based inference (as proposed by the baseline XDVioDet~\cite{wu2020not}).
Results on two visual corruptions (motion blur and Fog) and one audio Corruption (Random Dropout) provided in Fig.~\ref{fig:comparison_bs_prop_propdw} show that the naive averaging approach with our method outperforms the baseline in all cases, demonstrated the importance of shared space representation learning. Furthermore, the best results are achieved with the proposed dynamic weighting approach.



\subsection{Is Our Approach Plug-and-Play?}

The components of our approach, such as projection layer, shared representation learning, and dynamic weighting, can be plugged into existing multimodal anomaly detection methods, including XDVioDet~\cite{wu2020not}, HyperVD~\cite{zhou2024learning}, and RTFM~\cite{tian2021weakly}. These methods usually incorporate multimodal information by concatenating audio-visual modalities at feature-level input, making them prone to corrupted modalities during testing. As seen in Table \ref{tab:corruption_results}, when the components of our approach are added to XDVioDet and HyperVD, it outperforms the baseline in both cases of mixed corrupted modalities by a notable margin. In the case of RTFM as baseline, our method outperforms in the case of visual modality. However, in the case of corrupted audio modality, RTFM baseline performs better. This may be attributed to the projection layer used in our approach that upscale the audio features to match the dimensions of visual features. In contrast, RTFM by default uses the visual feature dimension of $2048$, whereas the dimension of added audio features is $128$.

\subsection{Does Multimodal Robustness Translate to Better Zero-shot Performance}

While the proposed shared space learning for multimodal data improves robustness against compromised modality, we explore its effectiveness in zero-shot setting. To this end, we utilize the models trained on XD-Violence dataset and evaluate on the test set of a benchmark anomaly detection dataset, UCF-crime \cite{sultani2018real}. As UCF-Crime is a visual-only dataset with no audio, this experiment poses two challenges: 1) Generalizability across datasets. 2) Missing audio data during testing. 
The results are summarized in Table \ref{tab:zero_shot}. As seen, our method outperformed RTFM and XDVioDet baselines as well as existing unsupervised video anomaly detection methods \cite{zaheer2022generative, al2024coarse,al2024collaborative}.

\subsection{Is Linear Projection Necessary?}
In this section, we explore the importance of the linear projection layer used in our approach. To this end, we replace it with zero-padding to match the embedding dimension of visual and audio modalities. The results are reported in Table \ref{tab:padding}. As seen, the padding demonstrates lower performance than even the baseline, whereas our design choice of using projection layer notably helps achieve better performance. This demonstrates that projection layer complements the proposed shared space representation learning for robust anomaly detection under multimodal setting, as proposed in our approach.

\begin{table}[t]
    \caption{AUC comparison on the test set of UCF-Crime dataset under zero-shot setting where the models are trained on XD-voilence dataset. Our approach demonstrates better zero-shot generalization, highlighting the importance of our proposed shared space representation learning. As seen, our approach also outperforms the existing unsupervised approaches trained and tested on UCF-crime~\cite{sultani2018real}.}
    \centering
    \footnotesize
    \begin{tabular}{l l c}
        \toprule
        \textbf{} & \textbf{Method} & \textbf{UCF-Crime} \\
        \midrule
        \multirow{3}{*}{Unsupervised} & Kim et al.~\cite{kim2021semi} & 52.00 \\
                                     & GCL~\cite{zaheer2022generative} & 65.32 \\
                                     & C2FPL~\cite{al2024coarse} & 65.85 \\
                                     & CLAP~\cite{al2024collaborative} & 67.74 \\
        \midrule
        \multirow{4}{*}{Zero-shot} & RTFM~\cite{tian2021weakly} (Baseline) & \underline{59.6} \\
                                   & RTFM (Ours) & \textbf{68.9 }\\
                                \cline{2-3}
                                  & XDVioDet~\cite{wu2020not} (Baseline) & \underline{63.3} \\
                                   & XDVioDet (Ours) & \textbf{68.4}\\

        \bottomrule
    \end{tabular}

    \label{tab:zero_shot}
\end{table}

\begin{table}[t]
\caption{Importance of linear projection compared to padding and concatenation (baseline) under different levels (0\% to 100\%) of an arbitrarily selected corruption case (pitch shift) for conciseness.}
\centering
\scriptsize
\begin{tabular}{l c c c c c c c}
\toprule
\textbf{Method} & \textbf{0\%} & \textbf{10\%} & \textbf{30\%} & \textbf{50\%} & \textbf{70\%} & \textbf{90\%} & \textbf{100\%} \\
\midrule
\textbf{Baseline} & 78.36 & 76.24 & 73.49 & 69.53 & 67.01 & 63.66 & 63.07 \\
\textbf{Padding}        & 76.21 & 73.46 & 70.52 & 64.30 & 60.15 & 56.83 & 55.86 \\
\textbf{Linear Proj.}   & 80.00 & 78.87 & 77.62 & 72.83 & 71.23 & 66.91 & 66.39 \\
\bottomrule
\end{tabular}

\label{tab:padding}
\end{table}

\subsection{Qualitative Results}

\begin{figure*}[t]
    \centering
    \includegraphics[width=0.99\textwidth]{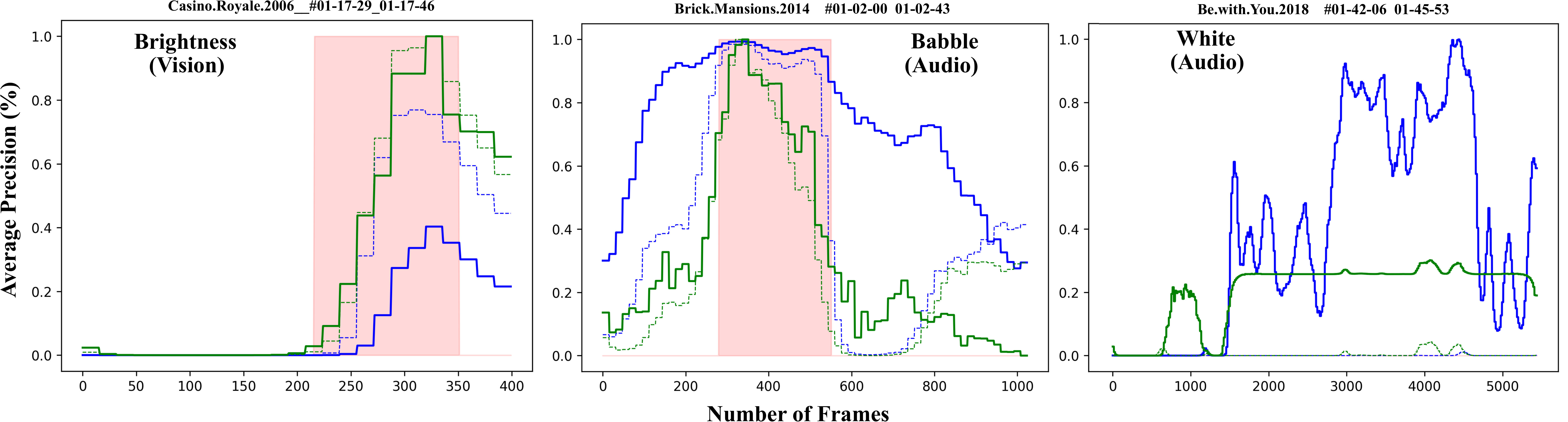} 
    \caption{Qualitative results of \our{} and the baseline on three videos taken from XD-Violence dataset. Blue represents baseline anomaly scores whereas green represents \our{} results. Dotter represents clean test samples whereas solid highlights corruption cases. As seen, our approach generally outputs comparable anomaly scores for clean and corruption cases. The baseline, while generating competitive anomaly scores for clean samples, demonstrates deteriorated performance when the input is corrupted. Red shaded area represents anomaly ground truth.}
    \label{fig:qualitative_subplots}
\end{figure*}

Fig. \ref{fig:qualitative_subplots} shows anomaly score plots of \our{} and the baseline on different videos highlighting cases with and without corruption. As seen, both \our{} and the baseline demonstrate reasonable anomaly scores when neither of the modalities is compromised. The performance of the baseline drops notably when modalities are corrupted, while \our{} retains the performance.

\section{Conclusion}
\label{sec:conclusion}
In this paper, we presented a comprehensive study to investigate the adverse effects of corrupted modalities on multimodal anomaly detection task. To address this, we introduced a novel dataset \our{}, to systematically evaluate the impact of audio and visual corruptions on anomaly detection performance. 
Moreover, we proposed a robust multimodal anomaly method that demonstrate resilience against corrupted and missing data. 
Extensive experiments on the XD-Violence dataset, across various corruption types and levels, demonstrated the effectiveness and robustness of our proposed method.
\our{} would be instrumental in evaluating anomaly detection methods under the settings more closer to the real-world scenarios.



\bibliographystyle{IEEEbib}
\bibliography{IEEEbib}

\newpage

 
\vspace{11pt}


\begin{IEEEbiography}[{\includegraphics[width=1in,height=1.2in,clip,keepaspectratio]{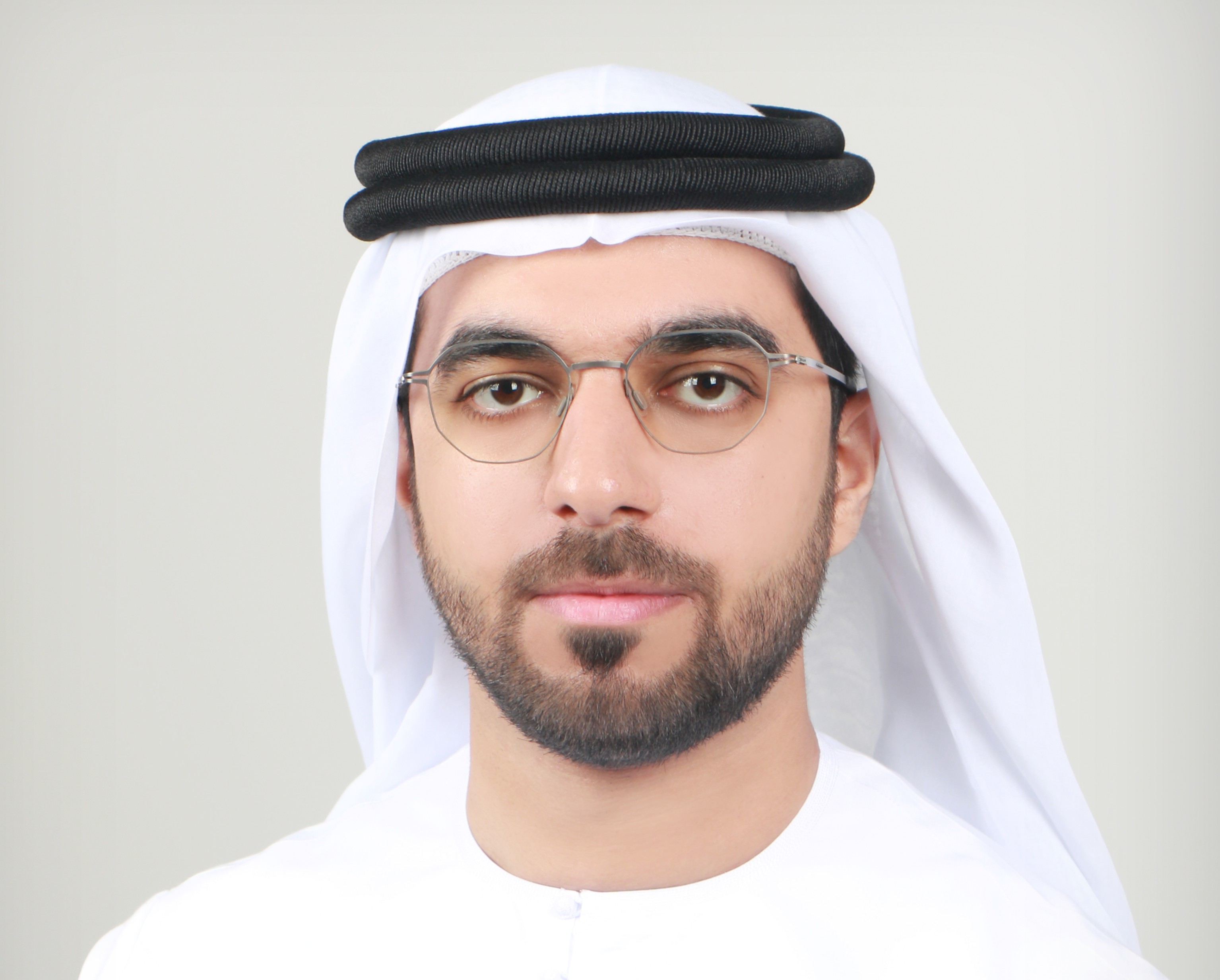}}]{Salem AlMarri} is a Ph.D. graduate from Mohamed Bin Zayed University of Artificial Intelligence, Abu Dhabi. He received his Masters degree in Electrical Engineering from Rochester's Institute of Technology. His thesis was on Utilizing Auxiliary Information for Weakly-Supervised Video Anomaly Detection.
\end{IEEEbiography}

\begin{IEEEbiography}[{\includegraphics[width=1in,height=1.2in,clip,keepaspectratio]{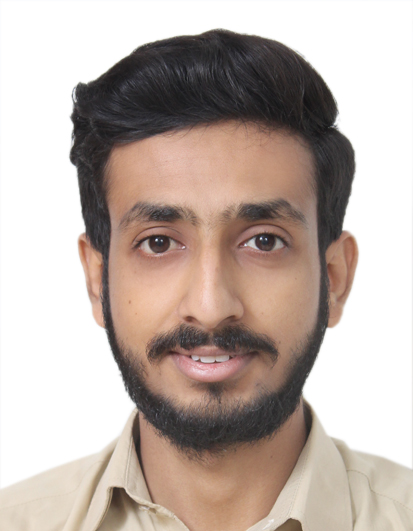}}]{Muhammad Irzam Liaqat} is a Ph.D. candidate in System Science at IMT School for Advanced Studies, Lucca, Italy. He received his master’s degree in computer science from the University of Engineering and Technology, Lahore, Pakistan. He held visiting researcher positions at Johannes Kepler University and at Mohamed bin Zayed University of Artificial Intelligence. His research interests include machine learning, deep learning, multimodal learning, computer vision, and medical imaging.
\end{IEEEbiography}

\begin{IEEEbiography}[{\includegraphics[width=1in,height=1.2in,clip,keepaspectratio]{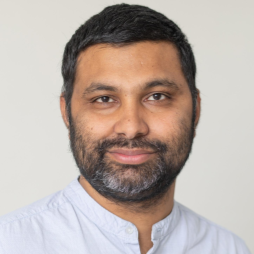}}]{Shah Nawaz} is an assistant professor at Johannes Kepler University Linz, Austria. He received bachelor degree in computer engineering from University of Engineering \& Technology, Taxila Pakistan and master degree in embedding systems from Technical University of Eindhoven, Netherlands, and PhD degree in computer science from University of Insubria, Italy. His research interests are focused on multimodal representation learning.
\end{IEEEbiography}

\begin{IEEEbiography}[{\includegraphics[width=1in,height=1.25in,clip,keepaspectratio]{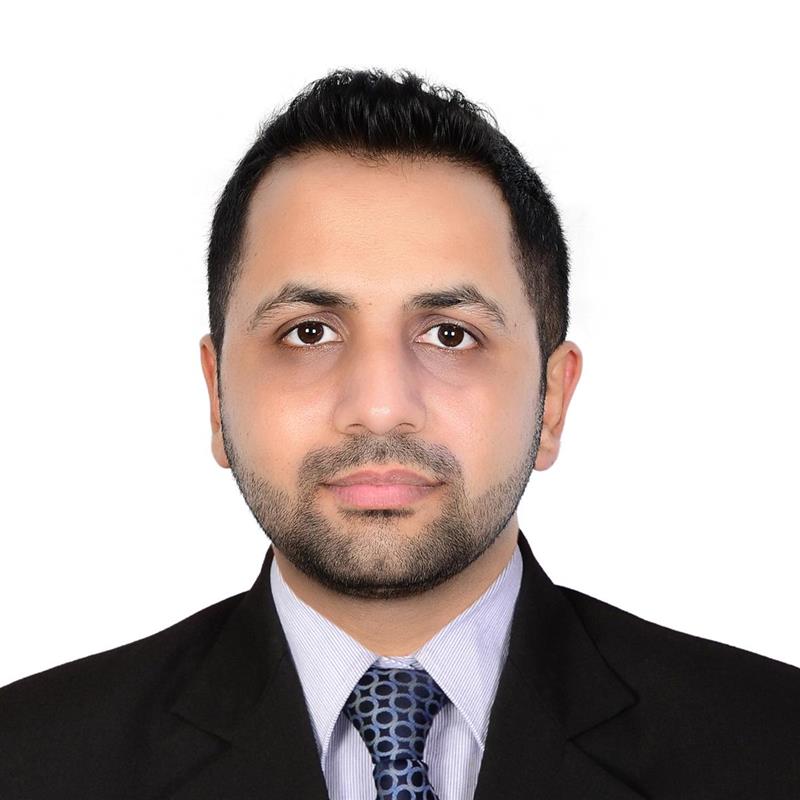}}]{Muhammad Zaigham Zaheer} received his PhD degree from the University of Science and Technology in 2022. He is currently associated with Mohamed bin Zayed University of Artificial Intelligence as a Research Scientist. Previously, he worked at the Electronics and Telecommunications Research Institute (ETRI) in South Korea, as a postdoctoral researcher.  His current research interests include vision language models and their applications, self-supervised learning, and unsupervised learning.
\end{IEEEbiography}

\begin{IEEEbiography}[{\includegraphics[width=1in,height=1.25in,clip,keepaspectratio]{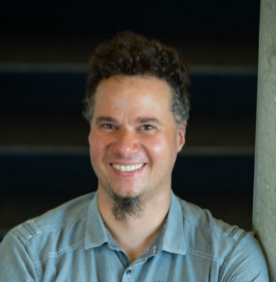}}]{Markus Schedl} is a full professor at the Johannes Kepler University Linz / Institute of Computational Perception, leading the Multimedia Mining and Search group. In addition, he is head of the Human-centered AI group at the Linz Institute of Technology (LIT) AI Lab. He graduated in Computer Science from the Vienna University of Technology and earned his Ph.D. from the Johannes Kepler University Linz. Markus further studied International Business Administration at the Vienna University of Economics and Business Administration as well as at the Handelshögskolan of the University of Gothenburg, which led to a Master's degree. His main research interests include recommender systems, information retrieval, natural language processing, multimedia, machine learning, and web mining. He (co-)authored more than 250 refereed articles in journals and conference proceedings as well as several book chapters. 
\end{IEEEbiography}

\begin{IEEEbiography}[{\includegraphics[width=1in,height=1.25in,clip,keepaspectratio]{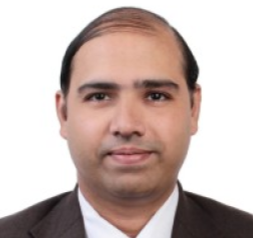}}]{Karthik Nandakumar} is an associate professor of computer vision at Mohamed bin Zayed University of Artificial Intelligence. His primary research interests include computer vision, machine learning, biometric recognition, applied cryptography, and blockchain. Prior to joining Mohamed bin Zayed University of Artificial Intelligence, Nandakumar was a research staff member at IBM Research – Singapore from 2014 to 2020 and a scientist at the Institute for Infocomm Research, A*STAR, Singapore from 2008 to 2014..
\end{IEEEbiography}




\vfill

\end{document}